\pdfoutput=1

\documentclass[11pt]{article}

\usepackage[final]{acl}

\usepackage{times}
\usepackage{latexsym}

\usepackage[T1]{fontenc}

\usepackage[utf8]{inputenc}

\usepackage{microtype}

\usepackage{inconsolata}

\usepackage[ruled]{algorithm2e}
\usepackage{graphicx}
\usepackage{multirow}
\usepackage{latexsym}
\usepackage{amsmath,amssymb,amsthm}
\usepackage{subfigure}
\usepackage{microtype}
\DeclareGraphicsExtensions{.pdf, .png}
\usepackage{epstopdf}

\usepackage{booktabs}
\usepackage{setspace}
\usepackage[normalem]{ulem}
\useunder{\uline}{\ul}{}
\usepackage{tikz}
\usepackage{bbding}
\usepackage{pifont}
\usepackage{wasysym}
\usepackage{paralist}
\usepackage{setspace}

\title{UniMEEC: Towards Unified Multimodal Emotion Recognition and Emotion Cause}

\author{Guimin Hu$^{\dagger}$, Zhihong Zhu$^{\spadesuit}$, Daniel Hershcovich$^{\dagger}$, Lijie Hu$^{\bigtriangleup}$, Hasti Seifi$^{\heartsuit}$, Jiayuan Xie$^{\diamond}$\\
  $^{\dagger}$University of Copenhagen \\
  $^{\spadesuit}$Peking University\\
  $^{\bigtriangleup}$King Abdullah University of Science and Technology\\
  $^{\heartsuit}$Arizona State University\\
  $^{\diamond}$The Hong Kong Polytechnic University\\
  \texttt{rice.hu.x@gmail.com}, \texttt{dh@di.ku.dk}, \texttt{hasti.seifi@asu.edu}}

\begin{document}
\maketitle
\begin{abstract}
Multimodal emotion recognition in conversation (MERC) and multimodal emotion-cause pair extraction (MECPE) have recently garnered significant attention. Emotions are the expression of affect or feelings; responses to specific events, or situations -- known as emotion causes. Both collectively explain the causality between human emotion and intents. However, existing works treat emotion recognition and emotion cause extraction as two individual problems, ignoring their natural causality. In this paper, we propose a \textbf{Uni}fied \textbf{M}ultimodal \textbf{E}motion recognition and \textbf{E}motion-\textbf{C}ause analysis framework ({\bf UniMEEC}) to explore the causality between emotion and emotion cause. Concretely, UniMEEC reformulates the MERC and MECPE tasks as mask prediction problems and unifies them with a causal prompt template. To differentiate the modal effects, UniMEEC proposes a multimodal causal prompt to probe the pre-trained knowledge specified to modality 
and implements cross-task and cross-modality interactions under task-oriented settings. Experiment results on four public benchmark datasets verify the model performance on MERC and MECPE tasks and achieve consistent improvements compared with the previous state-of-the-art methods.
\end{abstract}


\section{Introduction}
Recently, multimodal emotion recognition in conversations (MERC) and multimodal emotion-cause pair extraction (MECPE) have attracted increasing attention~\cite{Zhang_Xu_Lin_2021, Zhang_Xu_Lin_Lyu_2021, DBLP:conf/emnlp/HuLZ21,DBLP:journals/kbs/HuLZ21}. Both task play crucial roles in dialog systems, especially in empathetic response generation in a conversation~\cite{fu2023core,qian2023empathetic,tian2022empathetic,hu2024recent}.
\begin{figure}[t]
\centering
\includegraphics[width=\linewidth]{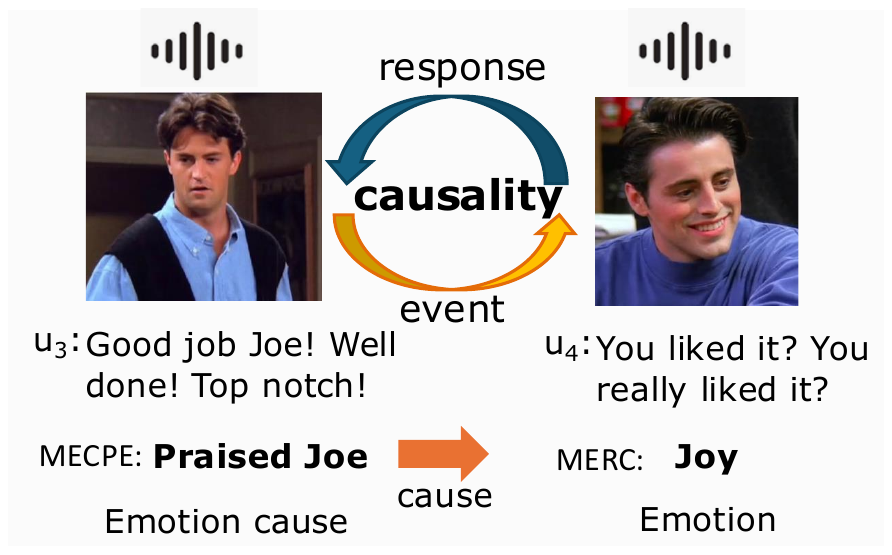}
\caption{Illustration of the causal inference between emotion and emotion cause, which unifies MECPE and MERC tasks. ``response'' denotes the speaker's reaction to the event and ``event'' denotes the event that triggers emotion.}
\label{fig:example}
\end{figure}
MERC detects the emotion category of each utterance in a conversation, while MECPE finds the reasons that trigger a certain emotion for the utterance. 
Both tasks are tightly related in practice and theory~\cite{baumeister1981can,dirven1997emotions,russell1990preschooler,lee2019emotion}. However, the existing works treat MERC and MECPE as two separate tasks and ignore their causality. On the one hand, emotions are responses to emotion causes (e.g., specific events)~\cite{marks1982theory,cabanac2002emotion}. On the other hand, emotion and its emotion causes are interdependent and mutually influential~\cite{russell1990preschooler,lee2019emotion}.
The two serve as reflections for each other and together provide a causal story of human behavior and intents.
Figure \ref{fig:example} illustrates the causal alignment between emotion category and emotion cause~\cite{baumeister1981can,dirven1997emotions}.

For example, the emotion causes of ``happiness'' generally are positive events, such as ``being praised''. Similarly, the emotion causes of ``sad'' generally are negative events, such as ``being criticized''. We view the mapping between the specific events (e.g., emotion cause) and response (e.g., emotion label) as the emotion-cause causality.
From the causal perspective, \citet{DBLP:journals/corr/abs-2404-11055} proposes the idea of causal prompts, which are prompts that describe the causal story behind the sentiment rating and reviews, further demonstrating that Pretrained Language Model (PLM) is able to be aware of the underlying causality. A natural question arises: \emph{How should we perform causality between emotions and their causes in a unified architecture? }


Recently, the unification of related but different tasks into a framework has achieved significant progress~\cite{DBLP:journals/corr/abs-2204-04637,DBLP:journals/corr/abs-2201-05966,DBLP:conf/aaai/ZhangMWJLY22}. For example, UniMSE~\cite{DBLP:conf/emnlp/HuLZLWL22} unifies emotion and sentiment into a single architecture to share complementary knowledge between them. Different from UniMSE which focuses on the unification of emotion and sentiment in a generative way, we propose a multimodal causal prompt to unify MERC and MECPE tasks, thereby capturing the causal nature between emotion and emotion cause.
In this paper, we propose a \textbf{Uni}fied \textbf{M}ultimodal \textbf{E}motion recognition and \textbf{E}motion-\textbf{C}ause pair extraction framework (UniMEEC) to explore the causality between emotion and emotion cause. As \citet{DBLP:journals/corr/abs-2404-11055} illustrated, PLM can capture the causal stories with the causal prompts. Starting from this perspective, UniMEEC reformulates MERC and MECPE as two mask prediction tasks and unifies the two tasks using a causal prompt, aiming to capture the understanding of PLM to emotion-cause causlity. In order to differentiate the modal effects, UniMEEC probes modal features from PLM using the multimodal causal prompt, and meanwhile, UniMEEC captures the emotion-specific, cause-specific, and utterance-specific contexts in a hierarchical way.
The main contributions are summarized as follows:



\begin{itemize}
    \item We propose a \textbf{Uni}fied \textbf{M}ultimodal \textbf{E}motion recognition and \textbf{E}motion \textbf{C}ause pair extraction framework ({\bf UniMEEC})\footnote{https://github.com/LeMei/causal-unimeec}, which uses the causal prompt to unify the MERC and MECPE tasks for causal relation between emotion and emotion cause.

    \item UniMEEC formalizes MERC and MEEC tasks into mask prediction problems and
    constructs the multimodal causal prompt to probe the knowledge from PLM. Meanwhile, UniMEEC proposes task-specific context aggregation to orderly capture the contexts oriented to specific tasks.

    \item Experimental results demonstrate that UniMEEC achieves a new state-of-the-art performance on MELD, IEMOCAP, ConvECPE and ECF datasets, further demonstrating the effectiveness of a unified causal framework for MERC and MECPE.

\end{itemize}

\section{Related Work}



\begin{figure*}[t]
\centerline{\includegraphics[width=0.9\textwidth]{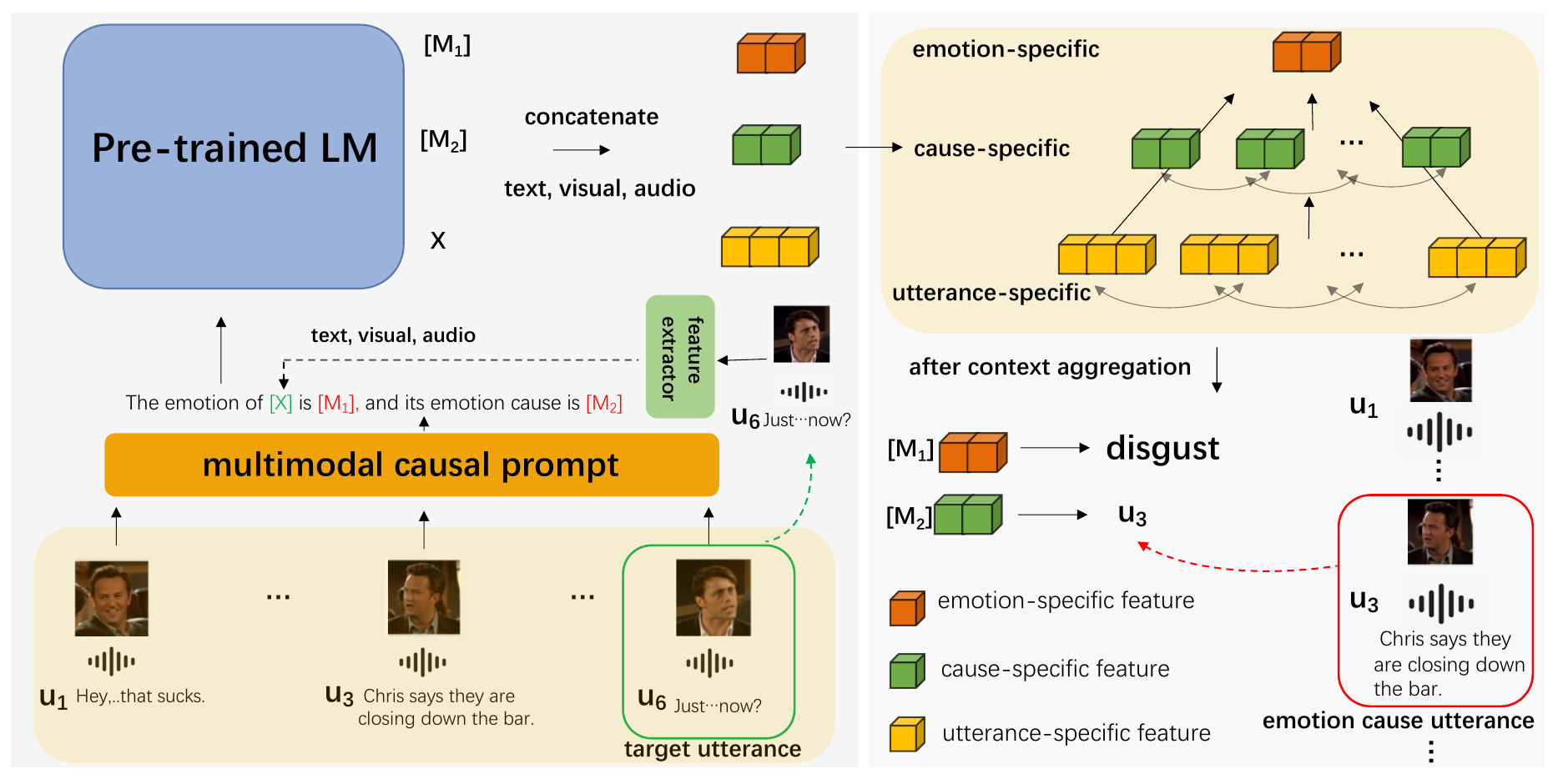}}
\caption{The overview of UniMEEC. The outputs ``disgust'' and ``$u_{3}$'' denote the emotion category and the emotion cause utterance ID of target utterance $u_{6}$, respectively.}
\label{fig:architecture}
\end{figure*}

\paragraph{Multimodal Emotion Recognition in Conversations (MERC)}
We categorize the works of MERC into three main groups: multimodal fusion, context-aware models, and external-knowledge models. The first group focuses on the fusion representation in which some works~\cite{DBLP:conf/icassp/HuHWJM22, DBLP:conf/acl/HuLZJ20, DBLP:journals/corr/abs-2205-02455} employed the graph neural networks to model the inter/intra dependencies of utterances information, and some works proposed cross-attention Transformer~\cite{DBLP:conf/nips/VaswaniSPUJGKP17} to model cross-modality interaction. Addressing context incorporation, \citet{DBLP:conf/emnlp/SunYF21,DBLP:conf/emnlp/Li00W21,DBLP:conf/emnlp/GhosalMPCG19} construct graph structures to represent contexts and further model inter-utterance dependencies, while~\citet{DBLP:conf/emnlp/Mao0WGL21} introduces the concept of emotion dynamics to capture context. In the last group, advanced MERC studies integrate external knowledge, employing techniques such as transfer learning~\cite{DBLP:journals/corr/abs-1910-04980,DBLP:journals/corr/abs-2108-11626}, commonsense knowledge~\cite{DBLP:conf/emnlp/GhosalMGMP20}, multi-task learning~\cite{DBLP:conf/naacl/AkhtarCGPEB19}, and external information ~\cite{DBLP:conf/acl/ZhuP0ZH20} to introduce more auxiliary information to help model understand conversation.

\paragraph{Multimodal Emotion-Cause Pair Extraction (MECPE)}
As more and more NLP tasks extend to the multimodal paradigm~\cite{zhu2024towards,li2024foodieqa,DBLP:zhu2024tfcd}, \citet{DBLP:journals/corr/abs-2110-08020} defined multimodal emotion-cause pair extraction (MECPE) and constructed Emotion-Cause-in-Friends (ECF) dataset based on MELD~\cite{DBLP:conf/acl/PoriaHMNCM19}. \citet{9926166} built an English conversational emotion-cause pair extraction multimodal dataset based on IEMOCAP~\cite{DBLP:journals/lre/BussoBLKMKCLN08}. With MECPE only emerging for a relatively short time, there are a few baseline methods in this field. Previous studie~\cite{DBLP:journals/corr/abs-2110-08020,9926166} integrated multimodal features to tackle the MECPE task based on the baselines of ECPE~\cite{DBLP:conf/acl/XiaD19}, overlooking the importance of inter-utterance context and multimodal fusion in understanding emotion cause.

\paragraph{Prompt-tuning}
Prompt-tuning~\cite{DBLP:conf/acl/LiL20,DBLP:journals/corr/abs-2110-07602,su2021transferability}, inspired by GPT-3~\cite{DBLP:conf/acl/DingQLCLJB23}, is a new paradigm to fine-tuning, particularly geared towards addressing few-shot scenarios. Recently, prompt-tuning has been widely used in addressing NLP tasks and achieved remarkable performances~\cite{zheng2022ueca,DBLP:journals/corr/abs-2109-08306,yang2023few,su2021transferability,sun2022tsgp}. The initial input $X$ undergoes modification through a template to form a textual string prompt $X'$ with unfilled slots. Subsequently, the language model is employed to probabilistically fill in the missing information, resulting in a final string $\hat{X}$ from which the model outputs $y$~\cite{liu2023pre}. The prompt template contains manual template engineering and automated template learning~\cite{liu2023pre}. The manual template is to manually create intuitive templates and the auto-prompt template~\cite{DBLP:conf/acl/LiL20,DBLP:journals/corr/abs-2110-07602,su2021transferability} includes discrete prompts, represented by actual text strings, and continuous prompts, described directly within the embedding space of the underlying language model. In this work, UniMEEC constructs causal prompts to unify MERC and MECPE, where causal prompt connects emotion and corresponding emotion cause to ensure the causal coherence.


\section{Methodology}

\subsection{Overall Architecture}
As shown in Figure \ref{fig:architecture}, UniMEEC is composed of multimodal causal prompt (MCP) and task-specific context aggregation (THC). Multimodal causal prompt template contains modality information [X], auxiliary prompt tokens $\text{P}_{(\cdot)}$, and mask tokens $\text{[M]}_\text{1}$ and $\text{[M]}_\text{2}$. We feed the causal template into PLM to encode [X], $\text{[M]}_\text{1}$ and $\text{[M]}_\text{2}$ into vectors. THC takes the emotion-specific, cause-specific, and utterance-specific representations as nodes and models their dependencies in the context window. Finally, UniMEEC predicts the emotion category and the position of cause utterance in a conversation based on the representations of $\text{[M]}_\text{1}$ and $\text{[M]}_\text{2}$ respectively.


\subsection{Task Formalization}
Given a multi-turn conversation $U=\{u_{1},u_{2},\cdots,u_{|U|}\}$, 
$U$ has $|U|$ utterances and each utterance $u_{i}=\{I^{t}_{i}, I^{a}_{i}, I^{v}_{i}\}$ contains three modalities, where $I^{m}_{i}, m\in \{t, a, v\}$ represent uni-modal feature extracted from video fragment $i$, and $\{t, a, v\}$ denote the three types of modalities—text, acoustic and visual, respectively. Multimodal emotion recognition (MERC) predicts the emotion category of $u_{i}$, and multimodal emotion-cause pair extraction (MECPE) aims to predict the corresponding cause utterance ID (e.g., ``$u_{1}$'', ``$u_{2}$'') for non-neutral utterance $u_{i}$. To unify MERC and MECPE, we formalize MERC and MECPE as two mask prediction problems in the causal prompt and leverage the language model to probabilistically fill the unfilled slots, thereby predicting the results of MERC and MECPE tasks respectively.

\subsection{Multimodal Causal Prompt (MCP)}
In order to differentiate the modal effects, we set causal prompt for each modality to probe the modality-specific features from PLM. Multimodal causal prompts share auxiliary prompt tokens in the prompt template, which enables inter-modality and inter-task semantic interaction in representation learning.

\subsubsection{Causal Prompt Construction}
We manually design the modality-specific prompt template, and it consists of a modal input [X], the emotion category slot $\text{[M]}_\text{1}$, the cause slot $\text{[M]}_\text{2}$ and auxiliary prompt part, where [X] is the slot filled with modal feature of target utterance, $\text{[M]}_\text{1}$ indicates the emotion category of target utterance, e.g., ``happy'' or ``sad'', and $\text{[M]}_\text{2}$ indicates the cause utterance ID of target utterance, e.g., ``$u_{1}$'', ``$u_{2}$''. $\text{[M]}_\text{1}$ and $\text{[M]}_\text{2}$ are unfilled answer slots and are separately predicted as the results of MERC and MECPE. Given text modality $I^{t}_{i},i\in \{1,\cdots,|U|\}$, we designed the causal prompt template like ``the emotion of utterance $I^{t}_{i}$ is $\text{[M]}_\text{1}$, and its emotion cause is $\text{[M]}_\text{2}$'' as text-specific prompt, where the textual strings ``For conversation'', ``the emotion category of'', ``is'', and ``the reason for this emotion is'' are auxiliary prompt parts. For audio-specific and vision-specific prompts, we replace the [X] part of the prompt with the acoustic and visual representations to construct audio-specific and vision-specific prompts, respectively. 

We use $X_{i,m}, X_{i,m}\in R^{l_{m} \times d_{m}}$ to represent the modal representation after modal alignment~\cite{DBLP:conf/acl/TsaiBLKMS19}, $l_{m}$ and $d_m$ are the sequence length and the representation dimension of modality $m$, respectively. Specifically, we obtain $X_{i,t}$ with the word embedding layer of the model and we processed raw acoustic input into numerical sequential vectors by librosa \footnote{\url{https://github.com/librosa/librosa}} to extract Mel-spectrogram as $X_{i,a}$. For vision modality, we use effecientNet~\cite{DBLP:conf/icml/TanL19} pre-trained (supervised) on VGGface \footnote{\url{https://www.robots.ox.ac.uk/~vgg/software/vgg\_face/}} and AFEW dataset to extract $X_{i,v}$.

\subsubsection{Causal Prompt Encoder}
We take Transformer-based model (e.g., BERT~\cite{devlin-etal-2019-bert}) as the backbone of the multimodal causal prompt. The stacked Transformer contains multiple Transformer layers, and each layer contains a self-attention module, FFN, and layer normalization~\cite{DBLP:journals/corr/BaKH16}. We take the former $N_{t}$ Transformer layers as the text-specific prompt encoder and take the latter $N_{a}$ and $N_{v}$ Transformer layers as the visual- and acoustic prompt encoders, respectively. First, text-specific prompt is fed into the text-specific prompt encoder to get the text-specific representations of [X], auxiliary prompt part, and $\text{[M]}_\text{1}$ and $\text{[M]}_\text{2}$, with the supervision of real ground answers of slots. After that, we obtain the text-specific prompt sequence, which contains the hidden states of $h_{P_{1,l_{1}}}$, $X_{i,t}$, $h_{P_{l_{2},l_{3}}}$, $h_{\text{[M]}_{1}}$, $h_{P_{l_{4},l_{5}}}$ and $h_{\text{[M]}_{2}}$, where $h_{(\cdot)}$ denotes the representation of token or token sequence, $h_{P_{1,l_{1}}}$, $h_{P_{l_{2},l_{3}}}$ and $h_{P_{l_{4},l_{5}}}$ denote the representations of auxiliary prompt parts.

Due to the dimensions and sequence lengths of audio and vision modalities being less than the dimensions and sequence length of text modality, we pad the audio and vision feature with zero to achieve consistency with the representation of text modality. We take $\hat{X}_{i,a}$ and $\hat{X}_{i,v}$ to represent audio and vision representations after padding, respectively. For audio-specific prompt, we replace [X] part of the prompt representation with $\hat{X}_{i,a}$. For vision-specific prompt, we replace [X] part of the prompt representation with $\hat{X}_{i,v}$ after $N_{t}$ Transformer layers. After that, we feed audio-specific and vision-specific prompts into $N_{a}$ and $N_{v}$ Transformer layers respectively. For (n-1)-th Transformer layer, the modality-specific prompt learning is given by:

\begin{small}
  \begin{align}
\begin{split}
    &P_{i,m}^{n-1} = [h_{P_{1,l_{1}}},X_{i,m}^{n-1},h_{P_{l_{2},l_{3}}},h_{\text{[M]}_{1}}^{m},h_{P_{l_{4},l_{5}}},h_{\text{[M]}_{2}}^{m}]\\
    &P_{i,m}^{n}=\text{Transformer}(P_{i,m}^{n-1},P_{i,m}^{n-1},P_{i,m}^{n-1})\\
    &X_{i,m}^{n} = P_{i,m}^{n}, m\in \{t,a,v\}
\end{split}
\end{align}  
\end{small}
where $P_{i,m}^{n-1}$ denotes the prompt representation of utterance $u_{i}$ under the modality $m$. Specifically, $P_{i,m}^{n-1}$ is composed by the hidden states of [X], $\text{[M]}_{1}$ $\text{[M]}_{2}$, and auxiliary prompt strings. $X_{i,t}^{0}=X_{i,t}$, $X_{i,a}^{0}=\hat{X}_{i,a}$, and $X_{i,v}^{0}=\hat{X}_{i,v}$. $[\cdot,\cdot]$ denotes the concatenation operation.

After the multimodal causal prompt, we obtain the modal fusion representations of mask tokens $\text{[M]}_{1}$ and $\text{[M]}_{2}$ via concatenation, respectively. Similarly, we obtain the fusion representation of $u_{i}$ via the concatenation of $X_{i,t}^{N_{t}}$, $X_{i,a}^{N_{a}}$ and $X_{i,v}^{N_{v}}$: 
\begin{small}
\begin{align}
\begin{split}
h_{\text{[M]}_{1}}^{f}&=[h_{\text{[M]}_{1}}^{t},h_{\text{[M]}_{1}}^{a},h_{\text{[M]}_{1}}^{v}]\\
h_{\text{[M]}_{2}}^{f}&=[h_{\text{[M]}_{2}}^{t},h_{\text{[M]}_{2}}^{a},h_{\text{[M]}_{2}}^{v}]\\
h_{u_{i}}^{f}&=[X_{i,t}^{N_{t}},X_{i,a}^{N_{a}},X_{i,v}^{N_{v}}]
\end{split}
\end{align}
\end{small}
where $X_{i,t}^{N_{t}}$, $X_{i,a}^{N_{a}}$ and $X_{i,v}^{N_{v}}$ are text, audio and video representations of $u_{i}$ encoded by $N_{t}$, $N_{a}$ and $N_{v}$ Transformer layers respectively.

\subsection{Task-specific Hierarchical Context (THC)}
The learned representations of $\text{[M]}_\text{1}$ (i.e., $h_{\text{[M]}_{1}}^{f}$) and $\text{[M]}_\text{2}$ (i.e., $h_{\text{[M]}_{2}}^{f}$) fail to capture the context information in a conversation, which inspires us to build a hierarchical context aggregation structure to control the direction of context aggregation in a conversation. In order to avoid the noise information in representation learning, we set the context windows for each utterance to incorporate the information around target utterance. 

\subsubsection{Hierarchical Graph Construction}
We construct a 3-level graph attention network (GAT) \cite{DBLP:conf/iclr/VelickovicCCRLB18} as the encoder of contexts, which includes top, middle, and bottom levels. Each level has a context window to focus on the local context of utterance. Formally, we define a graph $G = (V, E)$, $V$ and $E$ denote the node and edge sets respectively. We take the utterance-level representation $h_{u}$ as the bottom node, cause-specific token representation $h_{\text{[M]}_{2}}^{f}$ as the middle node, and the emotion-specific token representation $h_{\text{[M]}_{1}}^{f}$ as the top node. For the intra-level nodes, we set undirected edges for any two adjacent nodes in the context window of the same level. For the inter-level nodes, we set the undirected edges between the top nodes and middle nodes. In general, we set the directed edges from the bottom to the middle nodes in the context window, aiming to control the direction of the information flow among nodes.

Considering that graph $G$ contains multiple type node representations, we set five edge types respectively to model the dependency relations among different nodes. The former three edges are constructed between the slot nodes to slot nodes, i.e., 
$h_{\text{[M]}_\text{1}} \leftrightarrow h_{\text{[M]}_\text{1}}$, $h_{\text{[M]}_\text{1}}\leftrightarrow h_{\text{[M]}_\text{2}}$ and $h_{\text{[M]}_\text{2}} \leftrightarrow h_{\text{[M]}_\text{2}}$, which are represented with $t_{ee}$, $t_{ec}$ and $t_{cc}$ respectively.
The fourth edge type is constructed from utterance node to slot node, i.e., $h_{u}\leftrightarrow h_{\text{[M]}_\text{2}}$, represented by $t_{uc}$. The last is from utterance node to utterance node, i.e., $h_{u}\leftrightarrow h_{u}$, denoted by $t_{uu}$. The subscripts ``e'' and ``c'' in edge type represent $\text{[M]}_\text{1}$ and $\text{[M]}_\text{2}$, respectively, and ``u'' represents the utterance. For one edge type $t\in \{t_{ee},t_{ec},t_{cc},t_{uc},t_{uu}\}$, its adjacent matrix is given as:
\begin{align}
\begin{split}
a_{i,j}^{t} = \begin{cases}1 & j\in\{i-|w|,i+|w|\}\\0 & \text{otherwise}\end{cases}
\label{eq:adjacent}
\end{split}
\end{align} 
where $a_{i,j}^{t}\in A,A\in R^{V*V} $. $V$ denotes the number of utterances in a conversation. $|w|$ denotes the size of the context window. $i$ and $j$ represent the indexes of utterances in a conversation, and they are located on the same or adjacent levels of THC.

\subsubsection{Task-specific Context Aggregation}
We set a contextual window for each node at each level to ensure that the model only aggregates the node representations in its contextual window. This operation reduces the computational cost and avoids introducing noise to the representation learning. Given an utterance $u_{i}$, the prediction slots of emotion and emotion cause are $\text{[M]}_{i,1}$ and $\text{[M]}_{i,2}$ respectively. We aggregate the representation from the bottom to top levels in the graph, and the representations of bottom nodes are not updated by aggregating the representations of the top or middle nodes to them. For the bottom node $u_{i}$, its representation is aggregated by the bottom nodes in the context window:

\begin{small}
\begin{equation}
\begin{split}
&h_{u_{i}}^{n} = \operatorname{ReLU}\left(\sum_{j\in \mathcal{N}_{u_{i}}} a_{i,j}^{t_{uu}}W^{uu,n-1}h_{u_{j}}^{n-1}+b^{n-1}\right)
\end{split}
\end{equation} 
\end{small}
where $\mathcal{N}_{u_{i}}$ denotes the neighbor nodes of utterance $u_{i}$ and $h_{u_{j}}^{0}=h_{u_{j}}^{f}$. When the model comes to the middle node $\text{[M]}_{i,2}$, the representations is aggregated by the top and middle nodes in the context window, which is given by:

\begin{small}
\begin{equation}
\begin{split}
&h_{\text{[M]}_{i,2}}^{n} = \operatorname{ReLU}(\sum_{j\in \mathcal{N}_{\text{[M]}_{i,2}}} a_{i,j}^{t_{cc}}W^{cc,n-1}h_{\text{[M]}_{j,2}}^{n-1}\\
&+\sum_{j\in \mathcal{N}_{\text{[M]}_{i,1}}} a_{i,j}^{t_{ec}}W^{m_{ec},n-1}h_{\text{[M]}_{j,1}}^{n-1})\\
&+\sum_{j\in \mathcal{N}_{u_{i}}} a_{i,j}^{t_{uc}}W^{uc,n-1}h_{u_{j}}^{n-1}+b^{n-1})
\end{split}
\end{equation}  
\end{small}
where $\{\mathcal{N}_{\text{[M]}_{i,1}},\mathcal{N}_{\text{[M]}_{i,2}}\}$ denote the neighbor nodes of tokens $\text{[M]}_\text{1}$ and $\text{[M]}_\text{2}$ respectively. $h_{\text{[M]}_{j,1}}^{0}=h_{\text{[M]}_{j,1}}^{f}$, $h_{\text{[M]}_{j,2}}^{0}=h_{\text{[M]}_{j,2}}^{f}$. When the model comes to the top node $\text{[M]}_{i,1}$, its representation is aggregated by the top, and the middle nodes in the context window, which is given by:

\begin{small}
\begin{equation}
\begin{split}
&h_{\text{[M]}_{i,1}}^{n} = \operatorname{ReLU}(\sum_{j\in \mathcal{N}_{\text{[M]}_{i,1}}} a_{i,j}^{t_{ee}}W^{ee,n-1}h_{\text{[M]}_{j,1}}^{n-1}\\
&+\sum_{j\in \mathcal{N}_{\text{[M]}_{i,2}}} a_{i,j}^{t_{ec}}W^{ec,n-1}h_{\text{[M]}_{j,2}}^{n-1}+b^{n-1})
\end{split}
\end{equation} 
\end{small}
We stacked $N$ task-specific context aggregation modules and then use $h_{\text{[M]}_{i,1}}^{N}$ and $h_{\text{[M]}_{i,2}}^{N}$ as final representations of slots ${\text{[M]}_{i,1}}$ and ${\text{[M]}_{i,2}}$ respectively.

\subsection{Grounding Mask Predictions to MERC and MECPE}
We use $h_{\text{[M]}_{i,1}}^{N}$ to predict MERC task, i.e., the answers of slot $\text{[M]}_\text{1}$, and use $h_{\text{[M]}_{i,2}}^{N}$ to predict MECPE task, i.e.,  the answers of slot $\text{[M]}_\text{2}$. The predictions of $\text{[M]}_\text{1}$ (i.e., $\hat{y}_{i}^{e}$) and $\text{[M]}_\text{1}$ (i.e., $\hat{y}_{i}^{c}$) are given as respectively:

\begin{align}
\begin{split}
&\hat{y}_{i}^{e} = f(W^{e}h_{\text{[M]}_{i,1}}^{N} + b^{e})\\
&\hat{y}_{i}^{c} = f(W^{c}h_{\text{[M]}_{i,2}}^{N} + b^{c})
\end{split}
\end{align}
where $\{\hat{y}_{i}^{e},\hat{y}_{i}^{c}\}$ denote the prediction results for MERC and MECPE tasks, respectively. Based on the predictions, we use the sum of the cross-entropy losses of MERC and MECPE tasks as the objective loss of UniMEEC.

\section{Experiments}
\subsection{Datasets}
We conduct experiments on four publicly available benchmark datasets of MERC and MECPE. For MERC task, its benchmark datasets include multimodal emotionLines dataset (\textbf{MELD}) \cite{DBLP:conf/acl/PoriaHMNCM19}, interactive emotional dyadic motion capture database (\textbf{IEMOCAP}) \cite{DBLP:journals/lre/BussoBLKMKCLN08}. \textbf{IEMOCAP} consists of 7532 samples, and each sample is labeled with six emotions for emotion recognition, including happiness, sadness, anger, neutral, excitement, and frustration. \textbf{MELD} contains 13,707 video clips of multi-party conversations, with labels following Ekman’s six universal emotions, including joy, sadness, fear, angry, surprise and disgust.
\begin{table}[t]
\centering
            \fontsize{8}{10}\selectfont
\setlength{\tabcolsep}{4mm}{
\begin{tabular}{l|cccc}
\toprule
         Datasets & \textbf{Train} & \textbf{Valid} & \textbf{Test} & \textbf{All} \\
\midrule
MELD      & 9989    & 1108         & 2610    &   13707                        \\
IEMOCAP   & 5354    & 528           & 1650   & 7532               \\
ConvECPE   & 5303    & 486           & 1644   & 7433               \\
ECF   & 9457    & 1351           & 2701   & 13509         \\
\bottomrule
\end{tabular}}
\caption{The statistics of MELD, IEMOCAP, ConvECPE, and ECF.}
\label{tab:data}
\end{table}
For MECPE task, its benchmark datasets include \textbf{ConvECPE} \cite{9926166}, and emotion-cause-in-friends (\textbf{ECF}) \cite{DBLP:journals/corr/abs-2110-08020}. \textbf{ConvECPE} is a multimodal emotion cause dataset constructed based on IEMOCAP, in which each non-neutral utterance is labeled with the emotion cause. It contains 151 dialogues with 7,433 utterances. Similarly, \cite{DBLP:journals/corr/abs-2110-08020} annotated the emotion cause of each sample in MELD and then constructed multimodal emotion cause dataset ECF. \textbf{ECF} contains 1,344 conversations and 13,509 utterances. The detailed statistics of four datasets are shown in Table \ref{tab:data}.
For datasets IEMOCAP and MELD, we follow previous works \cite{li2021quantum,lu2020iterative}, and we use accuracy (ACC) and weighted F1 (WF1) as the evaluation metric for the MERC task. For datasets ECF and ConvECPE, we use precision (P), recall (R), and F1 as the evaluation metric for the MECPE task.
\begin{table*}[h]
\resizebox{\textwidth}{!}{
\begin{tabular}{l|ccccccc|cccccccc}
\toprule
\multirow{2}{*}{Methods} & \multicolumn{7}{c|}{IEMOCAP}                                              & \multicolumn{7}{c}{MELD}                \\
                        & {\bf Happiness} & {\bf Sadness} & {\bf Neutral} & {\bf Anger} & {
                        \bf Excitement} & {\bf Frustration} & {\bf WF1}  & {\bf Neutral} & {\bf Surprise} & {\bf Fear} & {\bf Sadness} & {\bf Joy}   & {\bf Disgust} & {\bf Angry} & {\bf WF1} \\
\midrule                    
BC-LSTM\cite{DBLP:conf/acl/PoriaCHMZM17}                 & 34.43     & 60.87   & 51.81   & 56.73 & 57.95      & 58.92      & 54.95 & 73.80    & 47.70     & 5.40  & 25.1    & 51.30  & 5.20     & 38.40  & 55.90                 \\
DialogueRNN\cite{DBLP:conf/aaai/MajumderPHMGC19}             & 33.18     & 78.80    & 59.21   & 65.28 & 71.86      & 58.91  & 62.75 & 76.23   & 49.59    & 0.00    & 26.33   & 54.55 & 0.81    & 46.76 & 58.73                \\
DialogueGCN\cite{DBLP:conf/emnlp/GhosalMPCG19}             & 51.87     & 76.76   & 56.76   & 62.26 & 72.71      & 58.04    & 63.16 & 76.02   & 46.37    & 0.98 & 24.32   & 53.62 & 1.22    & 43.03 & 57.52                \\
IterativeERC\cite{lu2020iterative}           & 53.17     & 77.19   & 61.31   & 61.45 & 69.23      & 60.92  & 64.37 & 77.52   & 53.65    & 3.31 & 23.62   & 56.63 & {19.38}   & 48.88& 60.72        \\
QMNN\cite{li2021quantum}                    & 39.71     & 68.30    & 55.29   & 62.58 & 66.71      & 62.19  & 59.88 & 77.00      & 49.76    & 0.00    & 16.50    & 52.08 & 0.00       & 43.17 & 58.00                   \\
MMGCN\cite{DBLP:conf/acl/HuLZJ20}                   & 42.34     & 78.67   & 61.73   & 69.00    & 74.33      & 62.32 & 66.22 &    -     &     -     &   -   &   -      &    -   &   -      &   -    & 58.65                \\
MM-DFN\cite{DBLP:conf/icassp/HuHWJM22}                   & 42.22     & 78.98   & 66.42   & 69.77    & 75.56      & 66.33 & 68.18 &    77.76     &     50.69     & -  &22.93   &   54.78     &   -      &   47.82    & 58.65                \\
MVN\cite{ma2022multi}                     & 55.75     & 73.30    & 61.88   & 65.96 & 69.50       & 64.21  & 65.44 & 76.65   & 53.18    & 11.70 & 21.82   & 53.62 & 21.86   & 42.55  & 59.03  \\
UniMSE\cite{DBLP:conf/emnlp/HuLZLWL22} & -     & -    & -   & - & -       & -   &  70.66 & -   & -   & - & -   & - & -   & - & 65.51 \\
EmoCaps\cite{li2022emocaps} & {\bf \ul 71.91}     & {85.06}    & 64.48   & 68.99 & {\ul 78.41}       & 66.76   & 71.77 & 77.12   & {\ul 63.19}    & 3.03 & 42.52   & 57.50 & 7.69   & {\ul 57.54} & 64.00  \\
GA2MIF\cite{zheng2023facial} & 46.15     & 84.50    & {\ul 68.38}   &  {\ul 70.29} & 75.99       & 66.49 & 70.00 & 76.92   & 49.08    & - & 27.18   & 51.87 & -  & 48.52 &58.94  \\
FacialMMT-RoBERTa\cite{zheng2023facial} & -     & -    & -   & - & -       & -   &  - & 80.13    & 59.63    & 19.18   & 41.99 & {\ul 64.88}       & 18.18       & 56.00 &{66.58}  \\
MALN\cite{DBLP:journals/tcsv/RenHLLLL23} & 55.50     & 81.80    & 64.10   & 69.10 & 78.00       & {\ul 71.40}  & 70.80 & {\ul 82.00}     & 58.60    & 21.20   & {\ul 43.00} & 64.30      & 17.60       & 52.40 &{\ul 66.90}  \\
MultiEMO\cite{DBLP:conf/acl/ShiH23}& 65.77     & {\ul 85.49}    & 67.08   & 69.88 & 77.31      & 70.98   &  {\ul 72.84} & 79.95    & 60.98    & {\ul 29.67}   & 41.51 & 62.82      & {\ul 36.75}       & 54.41 &{66.74}  \\
UniMEEC (Ours)   & {69.52}     & {\bf 88.51}    & {\bf 69.74}   & {\bf 72.63} & {\bf 78.80}      & {\bf 72.98}        & {\bf 74.83} & {\bf 82.75}   & {\bf 64.28}    & {\bf 31.78} & {\bf 43.31}   & {\bf 66.91} & {\bf 37.72}   & {\bf 58.46}  &{\bf 68.96}  \\
\bottomrule
\end{tabular}}
\caption{Results on IEMOCAP and MELD datasets. The best results are highlighted in bold. The results with underline denote the previous SOTA performance.}
\label{tab:main_results}
\end{table*}

\begin{table}[t]
\resizebox{\linewidth}{!}{
\scriptsize
\centering
\begin{tabular}{c|cccc}
\toprule
\multirow{1}{*}{}&\multicolumn{2}{c}{IEMOCAP}& \multicolumn{2}{c}{MELD} \\
\multirow{1}{*}{}&ACC & WF1 & ACC & WF1 \\
\midrule
BART&73.59&74.46&74.69&68.84\\
T5&74.32&75.09&74.93&69.06\\
LLaMA&74.67&75.16&75.02&69.15\\
\bottomrule
\end{tabular}}
\caption{Experimental results on IEMOCAP and MELD datasets with BART, T5 and LLaMA as backbone.}
\label{tab:encoder_results}
\end{table}

\subsection{Baselines}
For MERC, the baselines can be grouped into three categories: 1)the methods focusing on emotion cues like \textbf{EmoCaps} \cite{li2022emocaps}, \textbf{FacialMMT-RoBERTa} \cite{zheng2023facial}, \textbf{MVN} \cite{li2021quantum}. These works aim to improve model performance by tracking emotional states in a conversation, and 
2)the methods fusing multimodal information like \textbf{QMNN} \cite{li2021quantum}, \textbf{GA2MIF} \cite{li2023ga2mif},\textbf{MALN}\cite{DBLP:journals/tcsv/RenHLLLL23}, \textbf{MultiEMO} \cite{DBLP:conf/acl/ShiH23}, and \textbf{UniMSE} \cite{DBLP:conf/emnlp/HuLZLWL22}. These works focus on better multimodal fusion, and 
3)the methods incorporating context information like \textbf{DialogueGCN} \cite{DBLP:conf/emnlp/GhosalMPCG19}, \textbf{MMGCN} \cite{DBLP:conf/acl/HuLZJ20}, \textbf{MM-DFN} \cite{DBLP:conf/icassp/HuHWJM22}, \textbf{BC-LSTM} \cite{DBLP:conf/acl/PoriaCHMZM17}, \textbf{DialogueRNN} \cite{DBLP:conf/aaai/MajumderPHMGC19} and \textbf{IterativeERC} \cite{lu2020iterative}.
These works aggregate the context to understand the whole conversation.


MECPE has a few baselines due to MECPE only emerging for a relatively short time. Most baselines address MECPE tasks based on two-step frameworks of emotion-cause pair extraction in text, like {\bf Joint-GCN} \cite{9926166}, {\bf Joint-Xatt}\cite{9926166} and {\bf Inter-EC}\cite{9926166}.
{\bf $\textbf{C}_{\textbf{Multi-Bernoulli}}$}\cite{DBLP:journals/corr/abs-2110-08020} carries out a binary decision for each relative position to determine the cause utterance. {\bf $\textbf{C}_{\textbf{Multinomial}}$} \cite{DBLP:journals/corr/abs-2110-08020} randomly selects a relative position from all relative positions as the feature to extract emotion-cause pair. We produce some typical multimodal methods based on their open source codes, including {\bf MuLT} \cite{DBLP:conf/acl/TsaiBLKMS19}, {\bf MMGCN} \cite{DBLP:conf/acl/HuLZJ20}, {\bf MMDFN} \cite{DBLP:conf/icassp/HuHWJM22}, {\bf UniMSE} \cite{DBLP:conf/emnlp/HuLZLWL22} and {\bf GA2MIF} \cite{li2023ga2mif}.

\subsection{Experimental Settings} 
We use pre-trained BERT as the encoder of multimodal causal prompt. ConvECPE and ECF are constructed based on IEMOCAP and MELD respectively, so we integrate the emotion and cause labels of IEMOCAP, MELD, ConvECPE and ECF to train the model. The batch size is 64, the learning rate for BERT fine-tuning is set at 3e-4, and the learning rate for UniMEEC is set to 0.0001. The hidden dimension of acoustic and visual representation is 64, the BERT embedding size is 768, and the fusion vector size is 768. We use the former 9 Transformer layers of BERT as the text-specific prompt encoder, the following 10th and 11th as the audio-specific prompt encoder, and the last Transformer layer of BERT as the video-specific prompt encoder. The THC module stacks two graph network layers, where the first layer has one attention head and the second layer has four attention heads. 

\subsection{Experimental Environment}
\label{sec:environment}
All experiments are conducted in the NVIDIA RTX A100. We take BERT as the Transformer-based model, which has 110M parameters, including 12 layers, 768 hidden dimensions, and 12 heads. We use the former $N_{t}=9$ Transformer layers as the text-specific encoder, use the following $N_{a}=2$ and $N_{v}=1$ Transformer layers as the audio-specific and video-specific encoders respectively. The value of $N_{t}$, $N_{a}$ and $N_{v}$ are determined by the model  performance on valid test. Furthermore, we employ a linear decay learning rate schedule with a warm-up strategy.
\subsection{Results of Emotion Recognition}
We compare UniMEEC with the baselines of MERC on IEMOCAP and MELD datasets, and the comparative results are shown in Table \ref{tab:main_results}. UniMEEC significantly outperforms SOTA in all metrics on IEMOCAP, and MELD, and improves WF1 scores of IEMOCAP and MELD by 1.99\% and 2.06\%, respectively. 

Recent methods like MultiEMO, MALN, and GA2MF achieve low performance in recognizing the label ``Happiness'' for the IEMOCAP dataset and recognizing the label ``Fear'' for the MELD dataset. The low performance is caused by the label imbalance of the benchmark. UniMEEC significantly improves the emotion recognition performance on most emotion categories for two datasets. On the one hand, the unified framework offers model auxiliary information, enhancing the interaction between emotion and emotion cause, thereby alleviating the label imbalance of the benchmark. On the other hand, UniMEEC unifies the annotated labels of MERC and MECPE tasks with a causal prompt, which probes the causal story between response (emotion) and event (emotion cause). 
In summary, UniMEEC consistently surpasses the state-of-the-art (SOTA) in most emotion category recognition on both datasets. These results indicate the superiority of UniMEEC to MERC and MECPE and illustrate the unified framework of modeling emotion-cause causality brings improvements to emotion recognition.

Furthermore, we explore the impact of different PLMs, i.e., BART~\cite{DBLP:conf/acl/LewisLGGMLSZ20}, T5~\cite{DBLP:journals/jmlr/RaffelSRLNMZLL20} and LLaMa~\cite{DBLP:journals/corr/abs-2302-13971} on UniMEEC performance. We report the result on IEMOCAP and MELD datasets when we take BART, T5 and LLaMA as the PLM of UniMEEC. The experimental results are shown in Table \ref{tab:encoder_results}.


\begin{table*}[!ht]
\centering
\resizebox{\linewidth}{!}{
\begin{tabular}{l|ccccccc}
\toprule
\multirow{2}{*}{Methods} & \multicolumn{3}{c}{{Cause Recognition}} & \multicolumn{3}{c}{Pair Extraction}\\
                  &{ P}     &{ R}     &{ F1}       &{ P}    &{ R}    &{ F1}&{WF1}\\   
\midrule
{$\text{E}_\text{True}$ + $\text{C}_\text{Multi-Bernoulli}$}\cite{DBLP:journals/corr/abs-2110-08020}&55.69&57.20&55.47&49.40&25.22&33.39&-\\
{$\text{E}_\text{True}$ +$\text{C}_\text{Multinomial}$}\cite{DBLP:journals/corr/abs-2110-08020}&57.21&56.38&56.85&49.33&25.18&33.34&-\\
MC-ECPE-2steps\cite{DBLP:journals/corr/abs-2110-08020}&{\ul 57.76}&56.71&{\ul 57.09}&{\ul 49.43}&53.76&{\ul 51.32}&30.00\\
MuLT*\cite{DBLP:conf/acl/TsaiBLKMS19} & 55.19        & 53.43     & 54.79    & 30.48     & 37.85  & 39.02 &- \\
MMGCN*\cite{DBLP:conf/acl/HuLZJ20} & 56.51        & 54.82     & 55.30    & 35.43     & 38.19  & 37.48&54.65\\
MM-DFN*\cite{DBLP:conf/icassp/HuHWJM22} & 54.28        & 56.35     & 55.17    & 37.90     & 39.08  & 38.10 &54.86\\
UniMSE*\cite{DBLP:conf/emnlp/HuLZLWL22} & 56.55       & 57.09     & 56.73    & 44.48     & 54.25  & 49.08 &56.37\\
GA2MIF*\cite{zheng2023facial} & 56.48        & {\ul 58.33}     & 56.67    & 46.15     & {\ul 54.26}  & {50.16}  &{\ul 57.33} \\ 
 UniMEEC(Ours) &{\bf 59.87}&{\bf 58.85}&{\bf 59.18}& {\bf 49.88} & {\bf 59.29}  & {\bf 54.61}& {\bf 63.67}\\
\bottomrule
\end{tabular}}
\caption{Results on ECF dataset. Cause recognition is to predict the location of cause utterance and pair extraction is to match the emotion utterance and cause utterance, and WF1 denotes the performance of emotion recognition. The baselines with * are reproduced with their open sources.}
\label{tab:ecf_results}
\end{table*}
\begin{table*}[!ht]
\scriptsize
\centering
\resizebox{\linewidth}{!}{
\begin{tabular}{l|ccccccc}
\toprule
\multirow{2}{*}{Methods} & \multicolumn{3}{c}{{Cause Recognition}} & \multicolumn{3}{c}{Pair Extraction}\\
                  &{ P}     &{ R}     &{ F1}       &{ P}    &{ R}    &{ F1}&{WF1}\\                
\midrule
\textit{Joint-GCN(Joint-EC)}\cite{9926166}        & 71.47        & 86.35     & 78.21           & 38.23     & 37.08  & 37.65 & -                                 \\
\textit{Joint-Xatt(Joint-EC)}\cite{9926166}         & 69.68        & {\ul 89.42}     & 78.33    & 38.23     & 37.08  & 37.65 & -                               \\
\textit{Inter-EC}\cite{9926166}        & 68.55        & 85.55     & 76.11   & 30.91     & 37.34  & 33.82 & -                               \\
MuLT*\cite{DBLP:conf/acl/TsaiBLKMS19} & 75.15        & 71.43     & 73.05    & 44.61     & {\ul \bf 52.59}  & {\ul 48.74} & -\\
MMGCN*\cite{DBLP:conf/acl/HuLZJ20} & 78.57        & 74.52     & 76.07     & 42.18     & 42.67  & 42.11 & 63.28  \\
MM-DFN*\cite{DBLP:conf/icassp/HuHWJM22} & 79.84        & 74.11     & 76.90    & {\ul 46.79}     & 50.36  & 48.50  & 65.51 \\
UniMSE*\cite{DBLP:conf/emnlp/HuLZLWL22} & 80.37        & 73.09     & 75.58    & 44.24     & 49.33  & 46.69  & {\ul 67.36} \\
GA2MIF*\cite{zheng2023facial} & {\ul 81.42}        & 75.36     & {\ul 78.71}    & {46.54}     & 48.59  & 47.40 & -  \\
UniMEEC(Ours)  &{\bf 87.21}       &{\bf 92.95}  & {\bf 89.88}     & {\bf 50.61} & {50.41}     & {\bf 50.83}  & {\bf 69.48} \\
\bottomrule
\end{tabular}}
\caption{Results on ConvECPE dataset. The baselines with italics indicate it only uses textual modality.}
\label{tab:conv_results}
\end{table*}

\subsection{Results of Emotion-Cause Pair Extraction}
The results of cause recognition, pair extraction, and emotion recognition on ECF and ConvECPE datasets are shown in Table \ref{tab:ecf_results} and Table \ref{tab:conv_results}, respectively. 
UniMEEC significantly outperforms SOTA in all metrics on ECF and most metrics on 
ConvECPE datasets. For the ECF dataset, UniMEEC improves metrics P, R, and F of cause recognition by 2.11\%, 0.52\%, and 2.09\%, respectively, and P, R, and F of pair recognition by 0.45\%, 5.03\%, and 3.29\% respectively. For the ConvECPE dataset, multimodal methods perform better than text-based ones. UniMEEC improves by at least 2\% on most metrics for cause recognition and pair extraction. Furthermore, we report the UniMEEC performance of the emotion recognition task on two datasets (see WF1 in Table \ref{tab:ecf_results} and Table \ref{tab:conv_results}), outperforming at least 5.34\% and 2.12\% improvements by the competitive baselines on ECF and ConvECPE, respectively. 

We summarize the improvements into two aspects: 1) UniMEEC achieves SOTA on emotion recognition, cause recognition, and emotion-cause pair extraction on the benchmarks of MERC and MECPE, and 2)UniMEEC significantly outperforms SOTA in most cases. The improvements illustrate jointly training emotion and emotion cause can benefit the two tasks, and the unified framework in modeling causality between emotion and emotion cause can bring prior knowledge to MERC and MECPE training.

\subsection{Ablation Study}
We conducted extensive ablation studies on IEMOCAP and MELD datasets and experimental results are shown in Table \ref{tab:ablation}.
First, we remove the MECPE part in the prompt template, and then train UniMEEC just using the emotion label as the supervision signal. The removal of MECPE from UniMEEC results in a performance drop by 3.57\% and 1.96\% on IEMOCAP and MELD respectively, demonstrating that jointly training MERC and MECPE can bring improvements for MERC tasks.

Then we remove one or two modalities from MCP by replacing MCP with unimodal and bimodal prompt templates, where unimodal and bimodal prompt templates denote the prompt template containing one and two modalities, respectively. 
We feed the unimodal and bimodal prompts into PLM and their performances significantly decline on two datasets. We can find that removing acoustic, visual, and textual modalities or one of them all leads to performance degradation, further demonstrating the effectiveness and necessity of multimodal prompt learning to model performance. For example, we eliminate acoustic, visual, and both modalities from the multimodal prompt template, resulting in performance degradation by 2.75\%, 1.96\%, and 3.56\%, respectively, on WF1 for IEMOCAP. Similarly, the performance also drops for the MELD dataset after removing acoustic, visual, and both.
For the context aggregation module, we first remove THC from the model, which leads to 1.99\% and 3.54\% drops on two datasets respectively. Next, we disorder the positions of utterance-specific, cause-specific, and emotion-specific nodes in the THC module, disrupting the hierarchical structure of context aggregation, which results in 1.79\% and 1.94\% drops on IEMOCAP and MELD respectively. Additionally,  It can be found that removing the restriction of the context window when we construct the edges between nodes leads to the drop in ACC and WF1 on two datasets. Overall, MCP and THC are necessary to improve model performance, and introducing MERC and MECPE into a unified framework can bring improvements. 



\begin{table}[t]
\resizebox{\linewidth}{!}{
\scriptsize
\centering
\begin{tabular}{cl|cccc}
\toprule
\multirow{2}{*}{}& &\multicolumn{2}{c}{IEMOCAP}& \multicolumn{2}{c}{MELD} \\
\multirow{2}{*}{} & &ACC & WF1 & ACC & WF1 \\
\midrule
\multirow{1}{*}{Task}&- w/o MECPE& 68.55 & 71.26 & 71.41 & 66.79 \\
\midrule
\multirow{1}{*}{}&- w/o MCP & 68.04 & 72.70 & 71.52 & 65.32 \\
\multirow{3}{*}{UPL}&\quad- w/o A, T&68.02&71.19 & 69.74 & 62.96 \\
&\quad- w/o A, V  &69.37&72.84 & 70.65 & 65.05 \\
&\quad- w/o T, V & 68.59&71.88& 69.86 & 63.24 \\
\multirow{3}{*}{BPL}&\quad - w/o A & 70.19 & 72.08 & 73.42 & 65.66 \\
&\quad- w/o V & 71.02 & 72.87 & 73.65 & 66.89 \\
&\quad- w/o T & 67.75 & 71.23 & 69.76 & 65.47 \\
\midrule
\multirow{2}{*}{Context}&- w/o THC& 69.16 & 72.84 & 71.09 & 65.28 \\
&\quad - w/o hierarchy& 69.97 & 73.04 & 71.76 & 66.81 \\
&\quad - w/o $|w|$& 57.38 & 59.14 & 58.62 & 56.30 \\
\midrule
\textbf{UniMEEC (Ours)}& & \textbf{73.67} & \textbf{74.83} & \textbf{74.85} & \textbf{68.75} \\
\bottomrule
\end{tabular}}
\caption{Ablation study of UniMEEC on IEMOCAP and MELD datasets. T, V and A represent textual, visual and acoustic modalities, respectively. UPL and BPL denotes unimodal and bimodal causal prompts, respectively. Hierarchy denotes the hierarchical structure of THC.}
\label{tab:ablation}
\end{table}

\section{Conclusion}
This paper presents a unified multimodal emotion recognition and emotion-cause analysis framework, which aims to explore the emotion-cause causality by jointly modeling multimodal emotion recognition and emotion-cause pair extraction. UniMEEC reformulates MERC and MECPE tasks as two mask prediction problems, tunes PLM via multimodal causal prompts specific to uni-modality, and aggregates task-specific context in a conversation. Experiments on IEMOCAP, MELD, ConvECPE, and ECF consistently gain significant improvements on most metrics compared to the previous SOTA, further demonstrating the effectiveness of UniMEEC in addressing MERC and MEPCE. 

\section*{Limitations}
Due to the dimensions and sequence lengths of audio and vision modalities being less than the dimensions and sequence length of text modality, UniMEEC pads the audio and vision feature with zero to achieve consistency with the representation of text modality. This operation might introduce some unnecessary information in fusion representation learning. Furthermore, UniMEEC is set up to detect emotion and emotion cause in multimodal scenarios, fails to effectively address MERC and MECPE in text, which will also be solved in our future work. 

\section*{Ethics Statement}
The data used in this study are all open-source data for research purposes. While making machines understand human emotions and behaviors sounds appealing, it could be applied to emotional companion robots or intelligent customer service. However, even in simple multi-class emotion recognition , the proposed method can achieve only 74\% and 68\% in accuracy on IEMOCAP and MELD respectively, which is far from usable in real-world application.

\section*{Acknowledgement}
This work was supported by research grants from VILLUM FONDEN (VIL50296) and the National Science Foundation (\#2339707).

\bibliography{acl_latex}

\newpage
\appendix


\begin{table*}[t]
\resizebox{\linewidth}{!}{
\scriptsize
\centering
\begin{tabular}{cl|ccccccc}
\toprule
\multirow{2}{*}{}& &\multicolumn{3}{c}{{Cause Recognition}} & \multicolumn{3}{c}{Pair Extraction} \\
\multirow{2}{*}{} & &{ P}     &{ R}     &{ F1}       &{ P}    &{ R}    &{ F1}&{WF1}\\ 
\midrule
\multirow{2}{*}{Task}&- w/o MECPE& 56.16&54.39&54.64&48.35&58.68&52.41&60.78\\
&-w/o MCP&56.24&56.28&56.75&46.16&56.57&53.45&61.63\\
\midrule
\multirow{3}{*}{UPL}&-w/o A,T&56.25&56.41&56.09&46.09&56.47&53.72&61.41\\
&-w/o A,V&58.39&58.54&58.51&48.47&58.36&53.09&62.53\\
&-w/o T,V&56.43&56.77&56.25&46.14&56.54&53.82&61.29\\
\midrule
\multirow{3}{*}{BPL}&-w/o A&59.21&59.47&59.61&48.38&59.06&54.64&63.57\\
&-w/o V&59.46&59.63&59.62&48.54&58.32&54.07&63.75\\
&-w/o T&56.57&56.68&56.42&46.22&56.10&53.44&61.63\\
\midrule
\multirow{3}{*}{Context}&-w/o THC&57.32&56.36&55.19&47.41&57.26&53.43&62.94\\
&-w/o hierarchy&58.16&58.33&58.37&47.65&57.48&54.52&62.52\\
&-w/o |w|&56.61&56.63&56.56&46.63&56.41&52.47&62.43\\
\bottomrule
\end{tabular}}
\caption{Ablation study of UniMEEC on ECF dataset on cause recognition and pair extraction. T, V and A represent textual, visual and acoustic modalities, respectively. UPL and BPL denotes unimodal and bimodal causal prompts, respectively. Hierarchy denotes the hierarchical structure of THC.}
\label{appendix:ablation}
\end{table*}

\end{document}